\documentclass[runningheads]{llncs}
\usepackage[T1]{fontenc}
\usepackage{graphicx}
\usepackage{tikz}
\usetikzlibrary{decorations.pathreplacing}
\usepackage{amsmath}
\usepackage{amssymb}
\usepackage{todonotes}
\usepackage{subcaption}
\usepackage[nolist]{acronym}
\usepackage{booktabs}
\usepackage{multirow}
\usepackage{tabulary}
\usepackage{cite}
\usepackage{times}
\usepackage{amsmath,amssymb,amsfonts}
\usepackage{hyperref}

\usepackage{color}

\newboolean{useComments}
\setboolean{useComments}{FALSE}
\definecolor{dark_purple}{rgb}{0.1, 0.0, 0.4}
\definecolor{dark_green}{rgb}{0.0,0.2,0.5}
\definecolor{dark_red}{rgb}{0.85,0, 0}
\ifthenelse{\boolean{useComments}}{%
	\newcommand{\sr}[1]{\todo[inline,color=white!40,bordercolor=white]{\textcolor{blue!70!black!80}{\textbf{Sabine:}\textmd{\;#1}}}} 
	\newcommand{\jk}[1]{\todo[inline,color=white!40,bordercolor=white]{\textcolor{red}{\textbf{Jan:}\textmd{\;#1}}}}
	\newcommand{\vh}[1]{\todo[inline,color=white!40,bordercolor=white]{\textcolor{purple}{\textbf{Vahid:}\textmd{\;#1}}}}
}{%
	\newcommand{\sr}[1]{}
	\newcommand{\jk}[1]{}
	\newcommand{\vh}[1]{}
}

\begin{document}
	\begin{acronym}
		
		\acro{ood}[OOD]{Out-Of-Distribution}	
		\acro{id}[ID]{In-Distribution}
		\acro{nn}[NN]{Neural Network} 
		\acro{icad}[ICAD]{Inductive Conformal Anomaly Detection}
		\acro{fgsm}[FGSM]{Fast Gradient Sign Method}
		
	\end{acronym}
	\title{Runtime Monitoring for Out-of-Distribution Detection in Object Detection Neural Networks \thanks{This project has received funding from the European Union's Horizon 2020 Hi-Drive project under grant agreement No. 101006664 and the project Audi Verifiable AI.
		}
	}

	\titlerunning{Runtime Monitoring for OOD Detection in Object Detection Neural Networks}
	% FM is DOUBLE-BLIND
\author{Vahid Hashemi\inst{1}\and
	Jan K{\v{r}}et{\'\i}nsk{\`y} \inst{2} \and
	Sabine Rieder\inst{1,2} \and
	Jessica Schmidt\inst{1,3} }
\authorrunning{V. Hashemi et al.}
\institute{AUDI AG, Auto-Union-Stra\ss e 1, 85057 Ingolstadt, Germany \and
	Technical University of Munich, Germany \and	
	CISPA Helmholtz Center for Information Security, Stuhlsatzenhaus 5, 66123 Saarbrücken, Germany\\}
\maketitle              % typeset the header of the contribution
\begin{abstract}
	Runtime monitoring provides a more realistic and applicable alternative to verification in the setting of real neural networks used in industry. It is particularly useful for detecting out-of-distribution (OOD) inputs, for which the network was not trained and can yield erroneous results. We extend a runtime-monitoring approach previously proposed for classification networks to perception systems capable of identification and localization of multiple objects.
	Furthermore, we analyze its adequacy experimentally on different kinds of OOD settings, documenting the overall efficacy of our approach.
\end{abstract}
\section{Introduction}

\emph{\acp{nn}} can be trained to solve complex problems with very high accuracy. Consequently, there is a high demand to deploy them in various settings, many of which are also safety critical. In order to guarantee their safe operation, various verification techniques are being developed~\cite{DBLP:conf/date/ChengHBH20,Katz2019,Henriksen2020,Singh2019,robustnesscert,Zhang2018}. Unfortunately, despite the enormous effort, verification of NN of realistic industrial sizes is not within sight~\cite{Bak2021}. Therefore, more lightweight techniques, less depending on the size of the NN, are needed these days to provide some assurance of safety. In particular, \emph{runtime monitoring} replaces checking correctness universally on all inputs by following the current input only and raising an alarm, whenever the safety of operation might be violated.

Due to omnipresent abundance of data, NN can typically be trained well on these given inputs. However, they may work incorrectly particularly on inputs significantly different from the training data. Whenever such an \emph{\ac{ood}} input occurs, it is desirable to raise an alarm since there is much less trust in a correct decision of the NN on this input.
OOD inputs may be, for instance, pictures containing previously unseen objects or with noise stemming from the sensors or from an adversary.

In this paper, we provide a technique to efficiently detect such OOD inputs for the industrially relevant task of object detection, for which objects in an input image need to be localized and classified. We consider PolyYolo~\cite{polyyolo} as the object detection system of choice as it encompasses a very complex architecture like complex perception systems used in development of advanced driver assistance systems (ADAS) and autonomous driving functions. 
%kinds in an industrially relevant autonomous-driving application. 
Our approach builds upon a recent runtime-monitoring technique \cite{gaussbasedvahid} for efficient monitoring of classification networks.
As we consider object detection networks, the setting is technically different: the inputs are of a different type and, apart from classifying objects, their bounding boxes are to be produced. Even more importantly, the number of objects in the picture to be identified can now be more than 1 (often reaching dozens). As a result, questions arise how to apply the technique in this context, so that the efficiency and adequacy of the monitor is retained or even improved.

\emph{Our contribution} can be summarized as follows. We (i) propose how to extend the technique to this new setting (in Section \ref{sec:GaussExt}), (ii) improve and automate the detection mechanism (in Section \ref{sec:ICADExt}), and (iii) provide experiments on industrial benchmarks, concluding the efficacy of our approach (in Section \ref{sec:Experiments}).
In particular, our experiments focus on OOD due to pictures (i) from other sources, (ii) affected by random noise, e.g., from sensors, and (iii) affected by adversarial noise due to an FGSM attack \cite{fgsm}. 
On the methodological side, we leverage non-conformity measures to automate threshold setting for OOD detection.
Altogether, we extend the white-box monitoring approach \cite{gaussbasedvahid} to object detection systems more suited for real-world applications.
%, where activation values of selected neurons are tracked for training data

%Contributions:
%\begin{itemize}
%	\item Extension of a white-box monitoring approach to networks more suited for real world problems
%	\item Leveraging of non-conformity measures for automated threshold setting for OOD detection
%\end{itemize}

\paragraph{Related Work}

In this paper we focus on \ac{ood} detection when considering the neural network as a white box. OOD detection based on the activation values of neurons observed at runtime is extensively exploited in the state of the art~\cite{robustactivationmonitoring,activationpattern,gaussbasedvahid,outsidethebox,iterativemonitoring,ooddetectiongrammatrices}. In particular, Hashemi et al.~\cite{gaussbasedvahid} calculate the class-specific expectation values of all layer's neurons based on training data to abstract the \ac{id} behavior of the network. On top of that, they calculate the activations' confidence interval per class. At runtime if the network predicts a class but the activation values are not within the class-specific confidence interval, the result is declared as OOD as it does not match the expected ID behavior represented by the interval. 
Sastry et al.~\cite{ooddetectiongrammatrices} also monitor the network's activations during training. With this information, they calculate class-specific Gram matrices allowing them to detect deviations between the values within the matrix and the predicted class during the execution. Henzinger et al.~\cite{outsidethebox} use interval abstraction~\cite{intervalabstraction} where for each neuron an interval set is built which includes the neuron's activation values recorded while executing the training dataset. They utilize these constructed abstractions to identify novel inputs at runtime. 
In a follow-up work, Lukina et al.~\cite{{iterativemonitoring}} calculated distance functions to quantitatively measure the discrepancy between novel and in-distribution samples. Other directions of work for \ac{ood} detection involve generative models to measure the distance between the original image and the generated sample or monitoring of the last layer, e.g., \cite{reliableood,Wang2021}. 

Hendrycks et al. present different benchmarks for OOD detection in multi-class, multi-label and segmentation settings and apply baseline methods~\cite{Hendrycks2019}. They show that the MaxLogit monitor works well on all those problems. However, it is not directly applicable to the problem of object detection as in the other settings either the image or each pixel separately is assigned to classes. In the case of object detection, some parts of the image cannot be assigned meaningfully.

While all of the above techniques focus on classification or segmentation networks, we are only aware of few other approaches focusing on object detection neural networks. Du et al. \cite{Du2022} introduced a method for monitoring object detection systems by distilling unknown \ac{ood} objects from the training data and then training the object detector from scratch in combination with an uncertainty regularization branch. Similarly, \cite{Du2021} train an uncertainty branch by artificially synthesizing outliers from the feature space of the \ac{nn}. Consequently, the tools are not applicable to the frozen graph of a trained model. Unfortunately, this restriction beats the purpose of using (and monitoring) a \emph{given} trained network.%

We refer the reader to~\cite{oodsurvey} for a detailed overview on other monitoring approaches. 

\section{Preliminaries}
\subsection{Neural Networks}
\acfp{nn} are learning components which are often applied to complex tasks especially when it is hard to directly find algorithmic solutions. Examples of such tasks are classification, where the type of object in an image should be predicted, and object detection. In the latter case, images can contain several different objects at different locations. The \ac{nn} identifies the different objects in the image, assigns them to classes and computes \textit{bounding boxes}, usually of rectangular form,  surrounding the object. 

In general, a \ac{nn} consists of several consecutive \textit{layers} $1,...,L$ containing computation units called \textit{neurons}. The neurons receive their input as a sum from weighted connections to neurons in the previous layer and apply a usually non-linear \textit{activation function} $\sigma$ to their input. The result of this computation is called the \textit{activation value} $h$ of the neuron. More formally, the behavior of a neuron $j$ in layer $l+1$ with activation function $\sigma^{l+1}$ and incoming weights $w_{ij}$ from neuron $i \in N_l$ from layer $l$ with neurons $N_l$ can be described as follows for an input $x$:
\begin{equation*}
	h_j(x) = \sigma^{l+1} (\sum_{i \in N_l} w_{ij}h_i(x))
\end{equation*}
The activation values for neurons at layer $1$, which is called the \textit{input layer}, are defined as the input $x$:
\begin{equation*}
	\vec{h}^{\,1} (x) = x
\end{equation*}

\begin{figure}
	\centering
	\begin{tikzpicture}[node distance=0.5cm,
		node/.style={circle, draw=black, fill=black!5, thick, minimum size=4mm},
		interval/.style={rectangle, draw=orange, fill=orange!5, thick},
		scale=0.7]
		
		\node[node] (i1) at (0, 0) {};
		\node[node] (i2) at (0, 1) {} ;
		\node[node] (i3) at (0, 2) {} ;
		\node[node] (i4) at (0, 3) {} ;
		
		\node[node] (h1) at (2, 0.5) {};
		\node[node] (h2) at (2, 1.5) {} ;
		\node[node] (h3) at (2, 2.5) {} ;
		
		\node[node] (h4) at (4, 0.5) {};
		\node[node] (h5) at (4, 1.5) {} ;
		\node[node] (h6) at (4, 2.5) {} ;
		
		\node[node] (o1) at (6, 0) {};
		\node[node] (o2) at (6, 1) {} ;
		\node[node] (o3) at (6, 2) {} ;
		\node[node] (o4) at (6, 3) {} ;
		
		\draw[->] (i1) -> (h1);
		\draw[->] (i2) -> (h1);
		\draw[->] (i3) -> (h1);
		\draw[->] (i4) -> (h1);
		\draw[->] (i1) -> (h2);
		\draw[->] (i2) -> (h2);
		\draw[->] (i3) -> (h2);
		\draw[->] (i4) -> (h2);
		\draw[->] (i1) -> (h3);
		\draw[->] (i2) -> (h3);
		\draw[->] (i3) -> (h3);
		\draw[->] (i4) -> (h3);
		
		\draw[->] (h1) -> (h4);
		\draw[->] (h2) -> (h4);
		\draw[->] (h3) -> (h4);
		\draw[->] (h1) -> (h5);
		\draw[->] (h2) -> (h5);
		\draw[->] (h3) -> (h5);
		\draw[->] (h1) -> (h6);
		\draw[->] (h2) -> (h6);
		\draw[->] (h3) -> (h6);
		
		\draw[->] (h4) -> (o1);
		\draw[->] (h5) -> (o1);
		\draw[->] (h6) -> (o1);
		\draw[->] (h4) -> (o2);
		\draw[->] (h5) -> (o2);
		\draw[->] (h6) -> (o2);
		\draw[->] (h4) -> (o3);
		\draw[->] (h5) -> (o3);
		\draw[->] (h6) -> (o3);
		\draw[->] (h4) -> (o4);
		\draw[->] (h5) -> (o4);
		\draw[->] (h6) -> (o4);
		
		\node[] (d1) at (0, -0.5) {input layer};
		\draw [decorate, thick, decoration={brace, raise=2mm, amplitude=8pt}] (h4.east) -- (h1.west);
		\node[] (d2) at (3, -0.5) {hidden layers};
		\node[] (d1) at (6, -0.5) {output layer};
	\end{tikzpicture}
	\caption{Architecture of a \ac{nn}}
	\label{nn-architecture}
\end{figure}
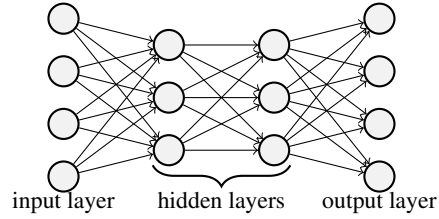

The last layer is the \textit{output layer}. The layers in between are called \textit{hidden layers}. An exemplary \ac{nn} is shown in Figure \ref{nn-architecture}.

The basic network architecture can be extended with different types of layers. Examples are convolutional, batch normalization and leaky ReLU layers. A convolutional layer takes its input as a 2- or 3-dimensional matrix and moves another matrix called the filter over the input. The input values are multiplied by the corresponding value in the filter to obtain the output. The goal of a batch normalization layer is to normalize the activation values of the neurons. Therefore, the mean and standard deviation are learned during training. During inference, the batch normalization layer behaves like a layer without an activation function as it only normalizes the activation values according to the learned parameters. The leaky ReLU layer takes only one input without weights and performs the following activation function: 
\begin{equation}
	\mathit{Leaky ReLU}(x) = 
	\begin{cases}
		x & \text{for } x > 0 \\
		0.01x & \text{otherwise}
	\end{cases}
\end{equation}
A more detailed introduction to \acp{nn} and different layer types can be found in \cite{nn}.

\subsection{Gaussian-Based White-Box Monitoring}
In \cite{gaussbasedvahid} Hashemi at al. introduced Gaussian-based \ac{ood} detection for a classification \ac{nn}. In this setting, the \ac{nn} is trained to assign an image to one of the classes in $C = \{c_1,...,c_{n_L}\}$. The underlying assumption is that neurons behave similar for objects of a particular class. Furthermore, neuron activation values are assumed to follow a Gaussian distribution. Therefore, the neuron activation values $h_i$ are recorded for each monitored neuron $i \in M$ for a set of monitored neurons $M$ and for each sample of the training data $X = \{x_1,..,x_m\}$ leading to a vector $\vec{r}^{\,i}$ with $r^i_j = h_i(x_j)$. The vector is then separated by class to $\vec{r}^{\,i}_{c_\star}$ for $c_\star \in C$. In the next step, the mean and standard deviation $\mu_{i, c_\star}, \sigma_{i, c_\star}$ are calculated for the neurons dependent on the classes. Due to assumption of a Gaussian distribution, $95\%$ of the samples are expected to fall into the range $[\mu_{i, c_\star} - k~\sigma_{i, c_\star}, ~\mu_{i, c_\star} + k~\sigma_{i, c_\star}]$ where $k$ is a value close to $2$. 

During inference, a new sample $x$ is fed into the \ac{nn}, a class $c_\star$ is predicted and the neuron activation values are recorded. The monitor checks if the activation values fall within the previously computed range of values. More formally: 
\begin{equation}
	\forall i\in M: h_i(x) \in [\mu_{i, c_\star} - k~\sigma_{i, c_\star},~ \mu_{i, c_\star} + k\sigma_{i, c_\star}]
\end{equation}

However, the paper \cite{gaussbasedvahid} showed that rarely the activation values of all neurons fall within the desired range.  Due to the selection of bounds for the interval to contain $95\%$ of the neuron activation values of the training data, even examples utilized to calculate the bounds may not fulfill the above condition. Therefore, the condition is weakened to only require a fixed percentage of neurons to be inside the bounds. This threshold was set manually in the paper with the goal of obtaining similar false alarm rates as Henzinger et al. \cite{outsidethebox}.

\subsection{Inductive Conformal Anomaly Detection}
In our work we leverage \ac{icad} which was introduced in \cite{laxhammer2015}. \ac{icad} extends conformal anomaly detection \cite{Laxhammar2014}. The idea is to predict if a new sample $x_{m+1}$ is similar to a given training set $X = \{x_1,..., x_m\}$. For this purpose, a nonconformity measure $A$ is introduced. This function takes as input the training set and a new sample for which to compute the nonconformity score and returns a real-valued measure of the distance of $x_{m+1}$ to the samples of $X$. Afterwards, the \textit{p-value} is calculated based on the nonconformity measure. The p-value for sample $x_{m+1}$ is calculated by

\begin{equation}
	p_{m+1} = \frac{|\{x_i \in X| A(X\setminus\{x_i\}, x_i) \geq A(X, x_{m+1})\}|}{|X|}.
\end{equation}

A low p-value hints to a non-conformal sample $x_{m+1}$. In general, this approach is inefficient as it requires the repeated computation of the nonconformity score for the entire training set $X$. An improvement was introduced in \cite{laxhammer2015}. The training set is split into a \textit{proper training set} $X_p = \{x_1,...,x_k\}$ and a \textit{calibration set} $X_c = \{x_{k+1}, ..., x_m\}$ with $k < m$. In the first step, the nonconformity measure $A$ is applied to samples of the calibration set based on the proper training set. For the new test sample $x_{m+1}$ the p-value is then computed in comparison to the calibration set:
\begin{equation}
	p_{m+1} = \frac{|\{x_i \in X_c | A(X_p, x_i) \geq A(X_p, x_{m+1})\}|}{|X_c|}
	\label{p-value}
\end{equation}

%\begin{itemize}
%	\item Deep neural networks
%	\item Poly-Yolo:
%	\begin{itemize}
	%		\item Why Poly-yolo: One-stage-detector, list the rest of detectors and name advantages of PolyYolo
	%		\item basic description of the architecture
	%	\end{itemize}
%	\item Non-conformity analysis (check VAE paper for information)
%	\item gaussian monitoring
%\end{itemize}

\section{Monitoring Algorithm}
In this paper we propose a monitoring algorithm which extends the Gaussian based monitoring from \cite{gaussbasedvahid} to object detection \acp{nn} and embeds it into the framework of \ac{icad}. 

\subsection{Extension to Object Detection Neural Networks} \label{sec:GaussExt}
The approach presented by Hashemi et al. \cite{gaussbasedvahid} relies on the distinction of images by different classes as  a separate interval for the neuron activation values is computed for each of the classes. However, images fed to an object detection network can contain several objects of different classes at different locations at the same time. When computing the intervals based on the classes contained in the images, one image could be relevant for several of those intervals. For example, an image containing a car and a pedestrian would contribute to the intervals for both classes. However, the pedestrian could only make up a small part of the input image leading to only a small fraction of neurons being influenced by the object. Consequently, neurons not related to the person are considered as relevant for the class intervals. Furthermore, the position of pedestrians throughout different images can shift and the neurons related to the pedestrian change accordingly. Consequently, the class related intervals would mostly consists of values from neurons that are not related to objects of the class. In addition, this approach increases the runtime at inference time. A previously unseen image would need to be checked against an interval for each class it contains an object of. In the worst case this could result in the total number of classes. As most of the values used for constructing the intervals are similar since they are not related to the particular object, the computations are also highly redundant.

To resolve both issues we discard the class information. This is supported by the observation that images are generally recorded in similar areas and therefore the general setting of a street is contained in all of them. The only changes are due to the objects and are locally bounded to their locations. The approach reduces the runtime to only one check per image and discards redundant computations. In total, we monitor the following condition discarding the class information: 
\begin{equation}
	\forall i\in M: h_i(x) \in [\mu_{i} - k~\sigma_{i}, ~\mu_{i} + k~\sigma_{i}]
\end{equation}

%\begin{itemize}
%	\item Extension to Multi-class setting by discarding the class information
%	\item Reason: In the last layers, only very tiny amount of neurons correspond to one anchor. Large portion of neurons not relevant for this output.
%	\item threshold setting by using the p-values
%\end{itemize}

\subsection{Embedding into the Framework of Inductive Conformal Anomaly Detection}\label{sec:ICADExt}
In the next step we improve the manual threshold setting from \cite{gaussbasedvahid} for the number of neurons that need to fall inside the expected interval. We propose to use \ac{icad} for this purpose. Therefore, we divide the training set into the proper training set $X_p$ and the calibration set $X_c$ and define the nonconformity measure $A$ to be the number of neurons falling \emph{outside} the range $[\mu_{i, p} - k~\sigma_{i, p}, ~\mu_{i, l, p} + k~\sigma_{i, p}]$ computed based on the proper training set $X_p$. We capture the number of neurons outside the interval rather than the ones inside as the nonconformity measure is expected to grow for OOD data. More formally with $M$ as the set of monitored neurons, usually all neurons of a particular layer and $\mu_{i, p}, ~\sigma_{i, p}$ the bounds computed as described in the last section based on the set $X_p$ as training set:
\begin{equation}
	A(X_p, x) = \frac{|\{i \in M | h_i(x) \notin [\mu_{i, p} - k~\sigma_{i, p}, ~\mu_{i, p} + k~\sigma_{i, p}]\}|}{|M|}
\end{equation}
\sr{Fixed a mistake in the formula}

Afterwards, the p-value is calculated as described in equation \ref{p-value}.  The threshold for the p-values is then set manually based on the requirements of the use case as there is a trade-off between the false alarm rate and the detection rate. For example, a high threshold for the p-vale leads to a low number of wrongly classified OOD examples, but the number of ID data classified as OOD will also rise as even some of the images from the calibration set are classified as OOD. Overall, the threshold setting is now closely related to the calibration set instead of the abstract metric of number of neurons inside the bounds.

\section{Experiments}\label{sec:Experiments}

\begin{figure}
	\centering
	\includegraphics[width=.5\linewidth]{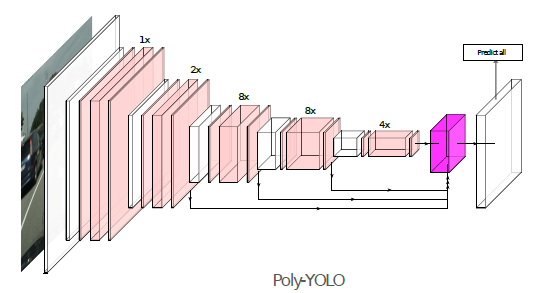}
	\caption{The image is taken from \cite{polyyolo} and shows the architecture of PolyYolo. White blocks represent convolutional sets, light pink indicates residual blocks with SE and dark pink shows the upsampling.}
	
	\label{fig:PolyArchitecture}
\end{figure}

Experiments were performed on PolyYolo \cite{polyyolo} which is based on the famous architecture  called YOLO (You Only Look Once)\cite{yolo}. YOLO was introduced in 2016 from Redmon et al. and afterwards continuously extended to improve the performance. For our work we decided to focus on PolyYolo\cite{polyyolo} as it improves YOLOv3 \cite{yolov3} while also reducing the size of the network. The architectue can be seen in Figure \ref{fig:PolyArchitecture}. PolyYolo consists of three main building blocks. A \emph{convolutional set} contains a convolutional layer and a batch normalization layer followed by leaky ReLU layer. A \emph{Squeeze-and-Excitation (SE) block}~\cite{Hu2018} contains a Global Average Pooling layer to reduce the size of each channel to 1 followed by a reshape layer, a dense layer, a leaky ReLU layer and a dense layer. The output of this sequence is meant to represent the importance of each channel compared to the others. Therefore, the last layer of the block multiplies the input with the result of the sequence to scale the input. The \emph{residual block with SE} then contains two consecutive convolutional sets followed by a SE block. The result is added to the input. The backbone of PolyYolo consists of several iterations of convolutional sets followed by residual blocks with SE as shown in figure \ref{fig:PolyArchitecture}. In between, there are three skip-connections to the neck. The neck uses upsampling to scale all results of the skip-connections to the same size and adds them up with intermediate convolutional sets. After all connections are added to one feature map, four convolutional sets are applied. The final layer is a convolutional layer. We monitored layers from the last convolutional set of the network as those are the last hidden layers and  Hashemi et al.~\cite{gaussbasedvahid} discovered that a monitor based on the last layers of a \ac{nn} lead to more accurate results. Namely we focus on the last batch normalization and leaky ReLU layer. As ID data we used Cityscapes \cite{cityscapes} which is the data set PolyYolo was trained on. 

\begin{figure}[!ht]
	\makebox[\linewidth][c]{
		\begin{subfigure}{.5\textwidth}
			\centering
			\includegraphics[width=.95\linewidth]{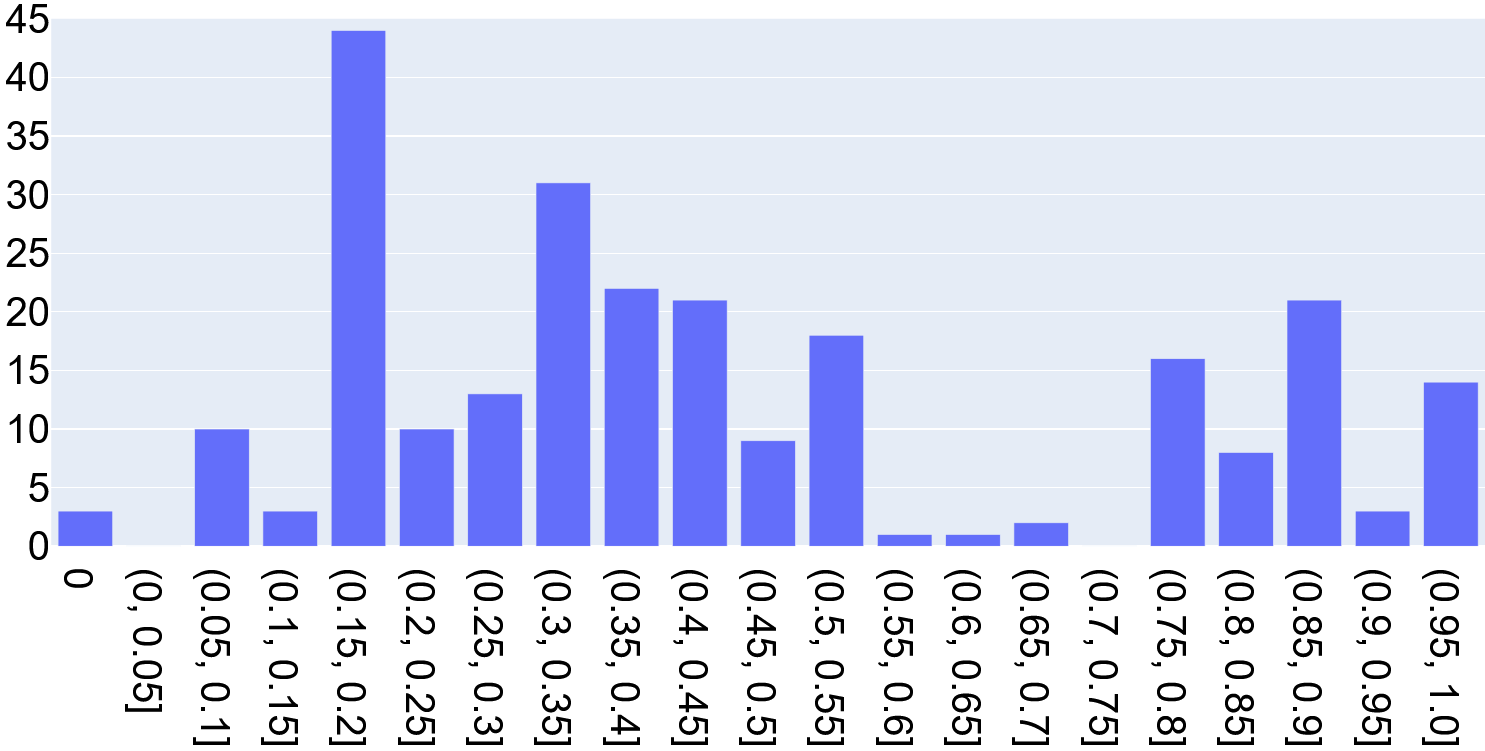}
			\caption{Last batch normalization layer with \newline $|X_c| = 20$ }
			\label{fig:sb10}
		\end{subfigure}%
		\begin{subfigure}{.5\textwidth}
			\centering
			\includegraphics[width=.95\linewidth]{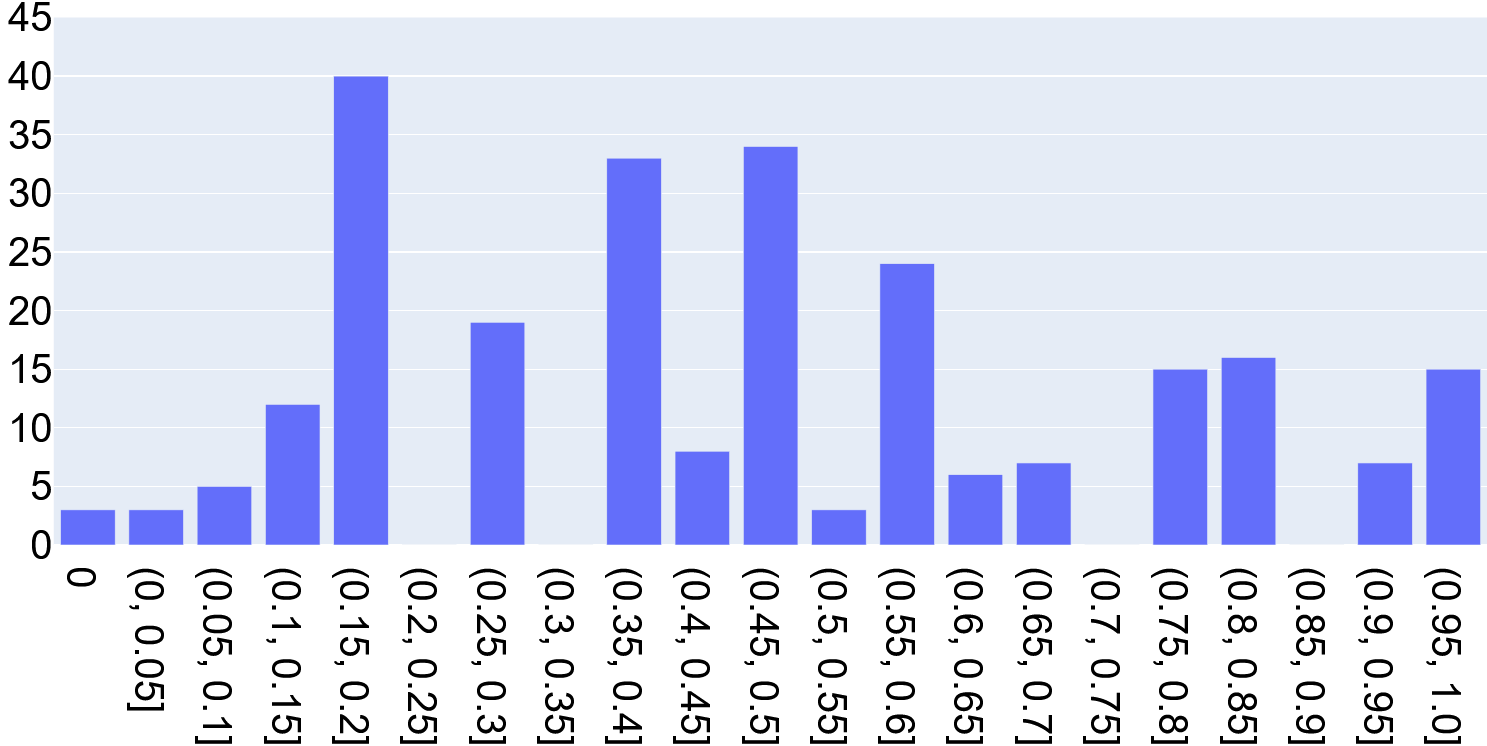}
			\caption{Last leaky ReLU layer with $|X_c| = 20$ \newline}
			\label{fig:sl10}
		\end{subfigure}
	}
	\makebox[\linewidth][c]{
		\begin{subfigure}{.5\textwidth}
			\centering
			\includegraphics[width=.95\linewidth]{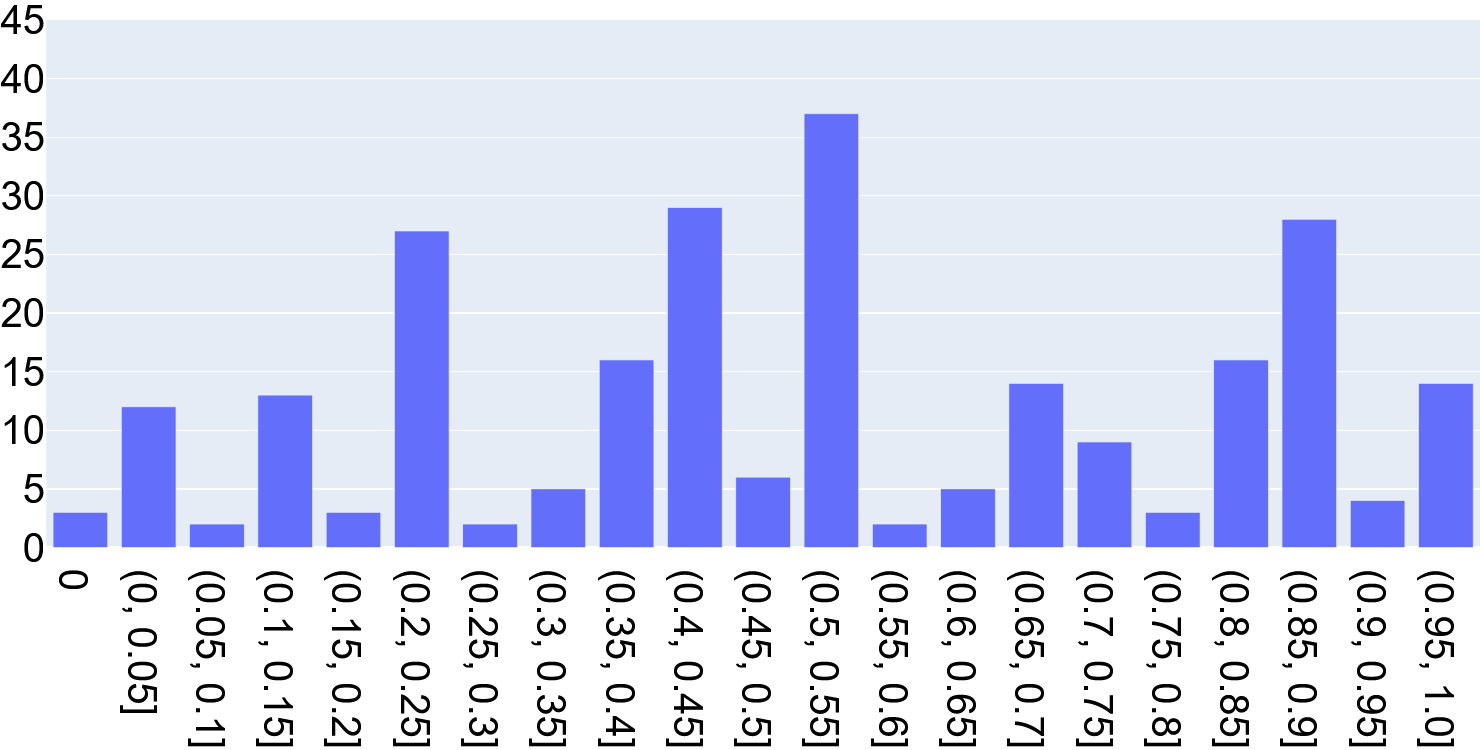}
			\caption{Last batch normalization layer with \newline $|X_c| = 60$ }
			\label{fig:sb50}
		\end{subfigure}%
		\begin{subfigure}{.5\textwidth}
			\centering
			\includegraphics[width=.95\linewidth]{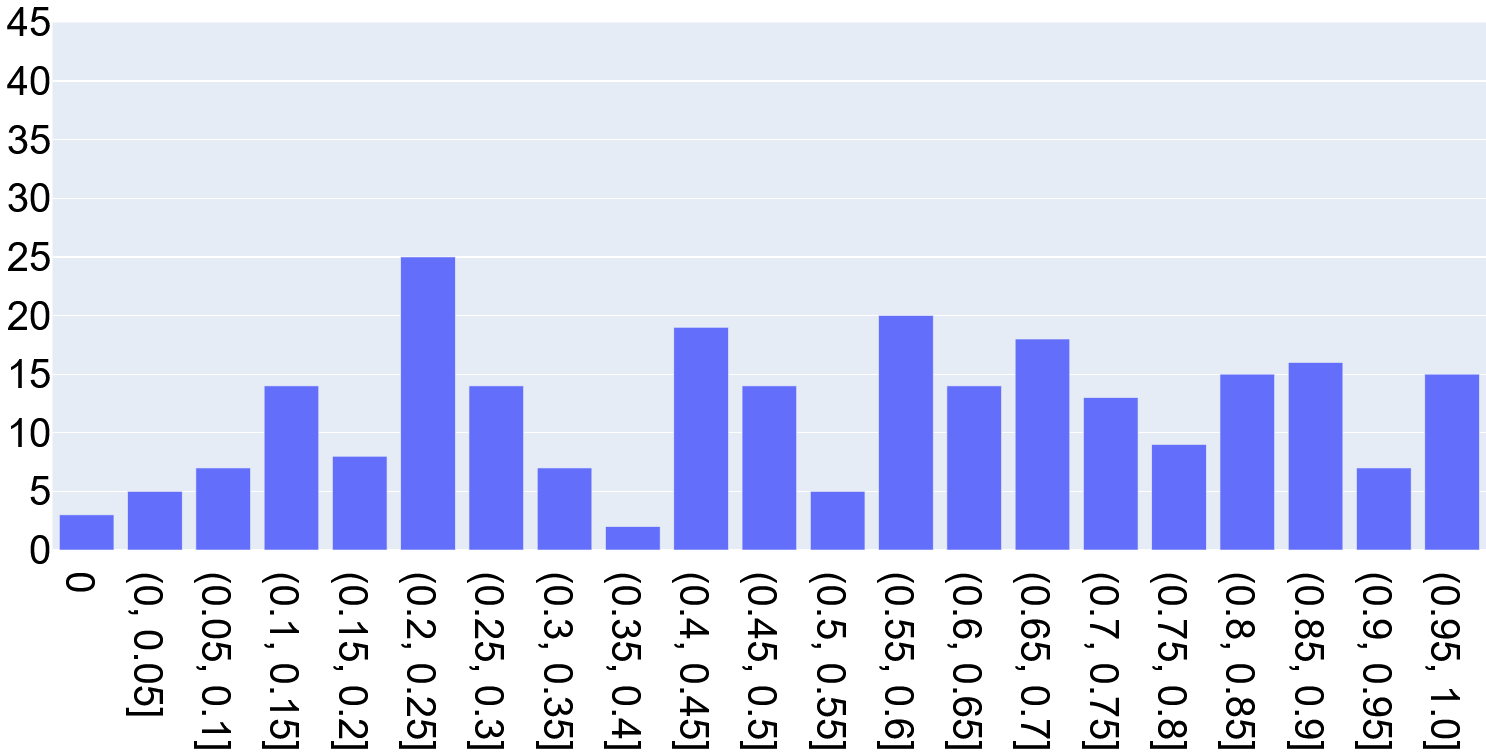}
			\caption{Last leaky ReLU layer with $|X_c| = 60$\newline }
			\label{fig:sl50}
		\end{subfigure}
	}
	\makebox[\linewidth][c]{
		\begin{subfigure}{.5\textwidth}
			\centering
			\includegraphics[width=.95\linewidth]{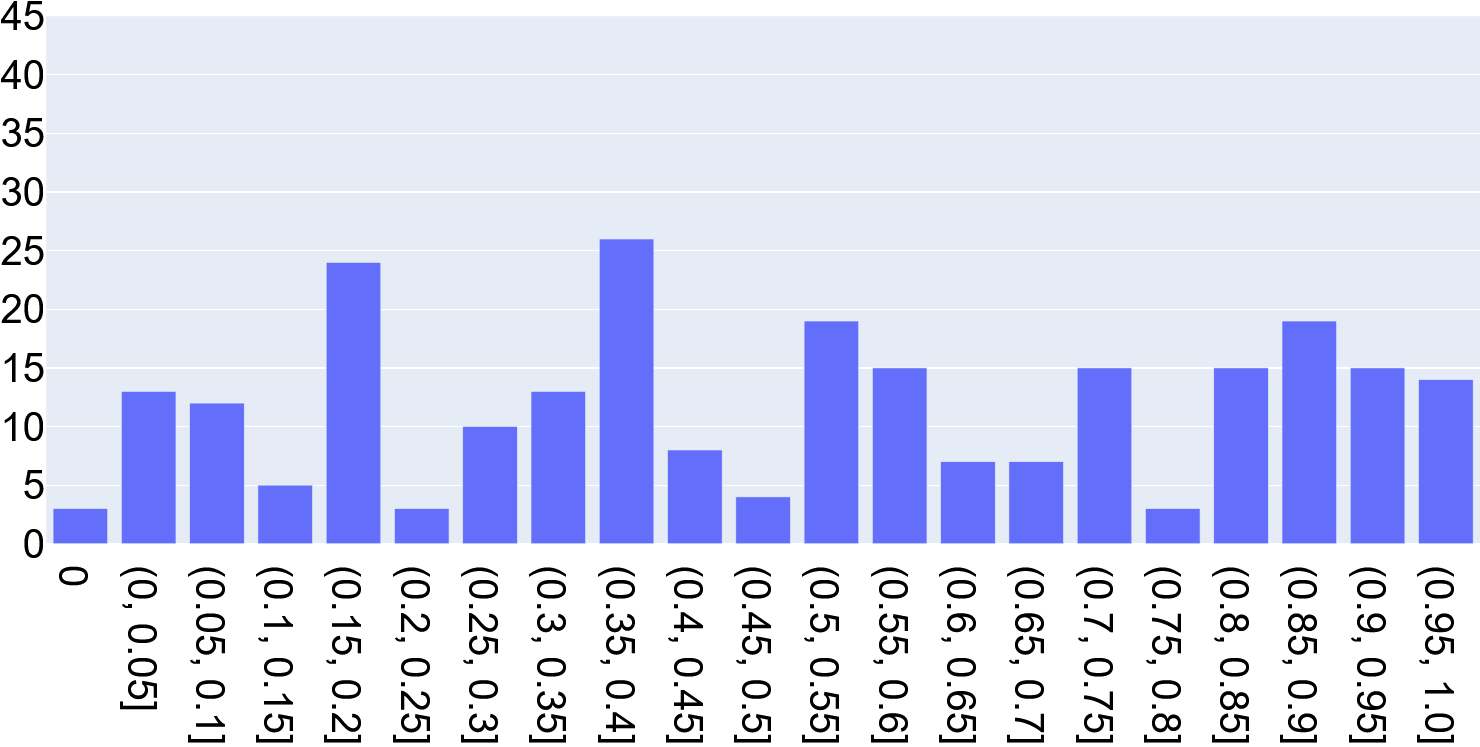}
			\caption{Last batch normalization layer with \newline $|X_c| = 100$ }
			\label{fig:sb100}
		\end{subfigure}%
		\begin{subfigure}{.5\textwidth}
			\centering
			\includegraphics[width=.95\linewidth]{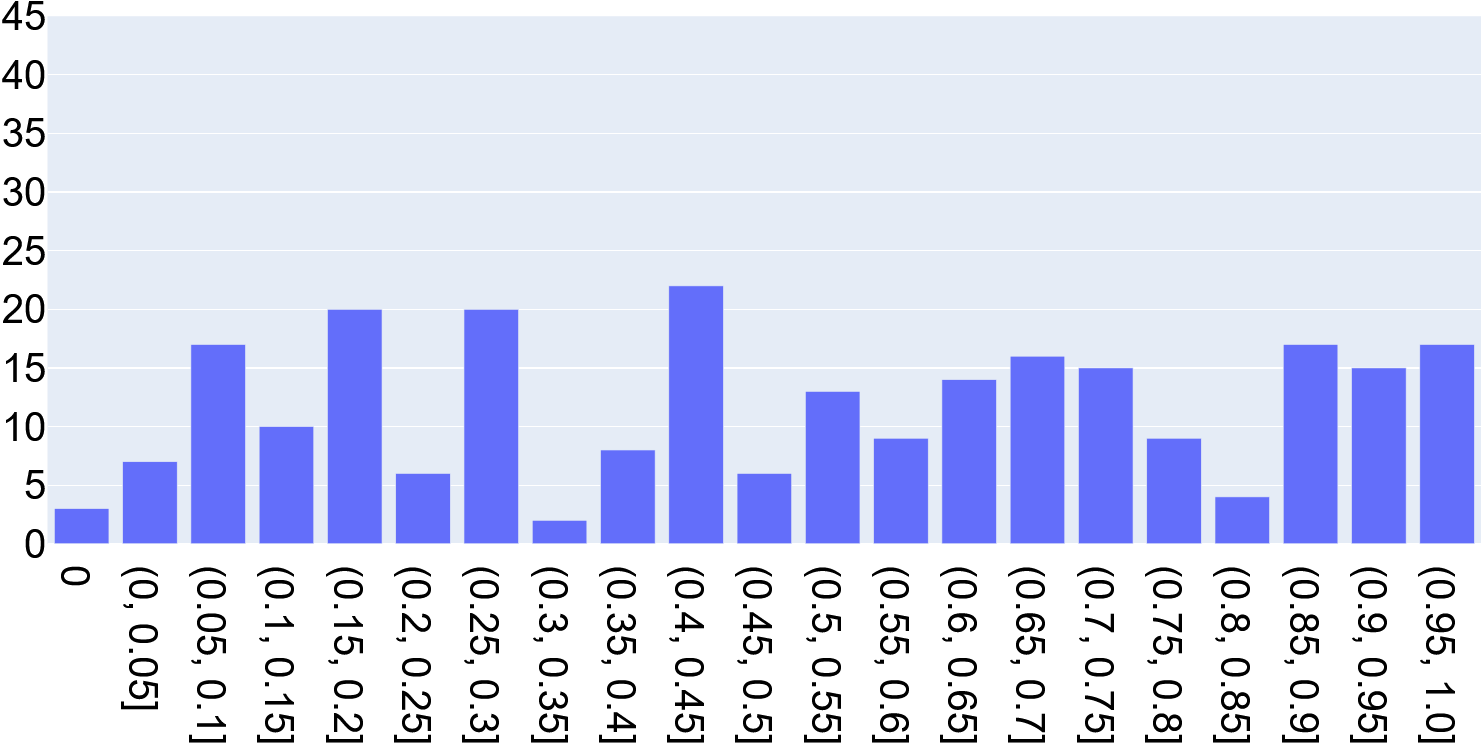}
			\caption{Last leaky ReLU layer with $|X_c| = 100$\newline }
			\label{fig:sl100}
		\end{subfigure}
	}
	\caption{The x-axis shows the range of p-value and the y-axis the number of images resulting in a p-value contained in the interval. The rows correspond to different sizes of calibration sets while the columns contain the monitored layers.}
	\label{fig:differentXc}
\end{figure}

We computed intervals for the neuron activation values based on 500 training images of the Cityscapes data set and the calibration set consists of 100 test images of Cityscapes. In a first step, we investigated the size of the calibration set. Figure \ref{fig:differentXc} shows the importance of including images with different features. The x-axis shows the interval of p-values considered for the bar while the y-axis shows the number of images resulting in a p-value within this interval. For a calibration set of size $20$, many samples obtain a p-value in the interval $(15, 20]$. For a large calibration set, the peaks in the graph are flattened. However, it is also noticeable that some elements of $X_c$ are of more importance to the test data than others resulting in peaks as they separate the test data. Small bars in the graph are the result of elements of $X_c$ that do not contribute a value for the nonconformity measure with huge difference to their neighbors. Therefore, samples from the test data that have a higher nonconformity score than these images also have a larger nonconformity score than other samples of $X_c$. A more advanced selection strategy for the calibration set could reduce this effect. To this end, we therefore fix the size of the calibration set to 100 images.

\begin{figure}
	\makebox[\linewidth][c]{
		\begin{subfigure}{.5\textwidth}
			\centering
			\includegraphics[width=.95\linewidth]{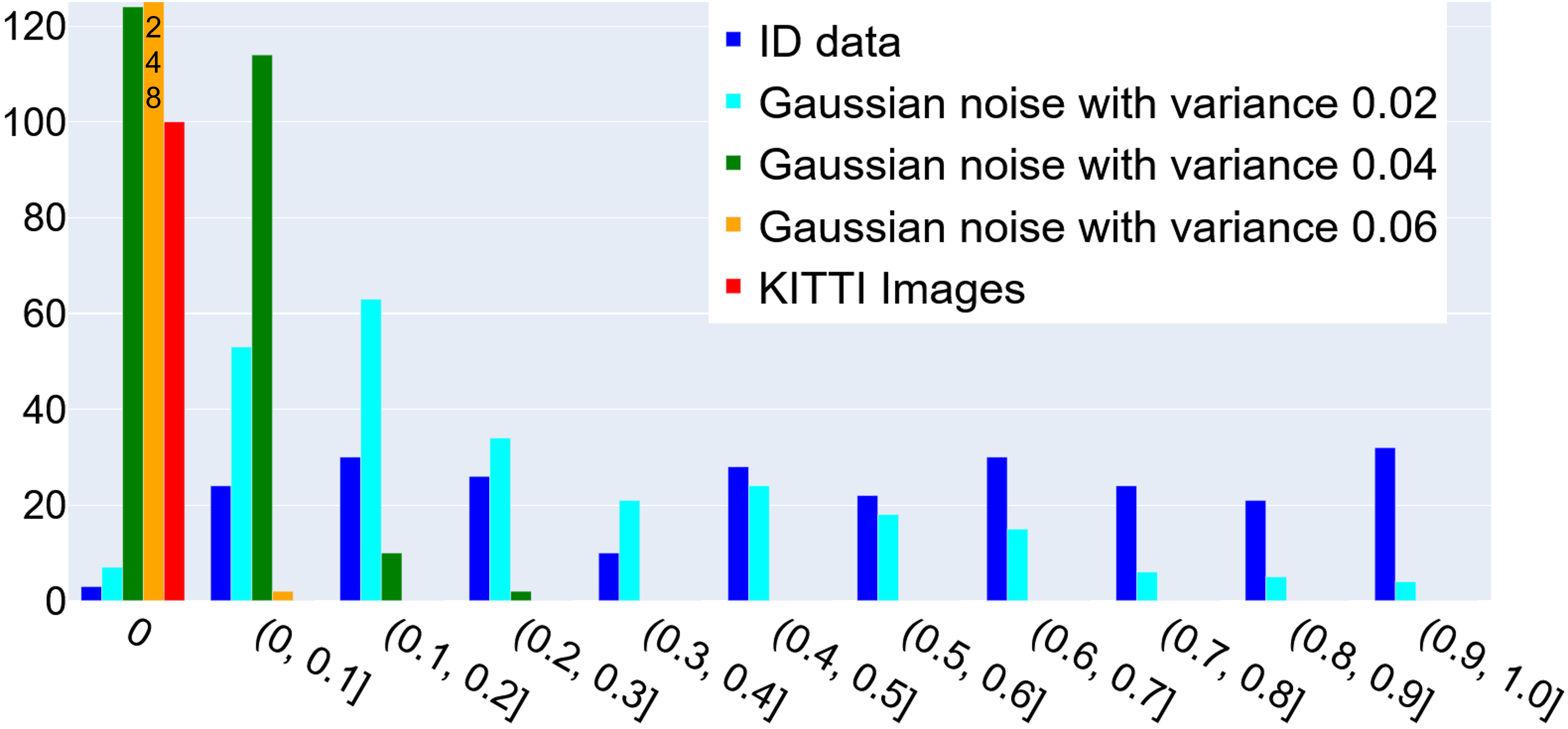}
			\caption{Last leaky ReLU layer}
			\label{fig:sfig1com}
		\end{subfigure}%
		\begin{subfigure}{.5\textwidth}
			\centering
			\includegraphics[width=.95\linewidth]{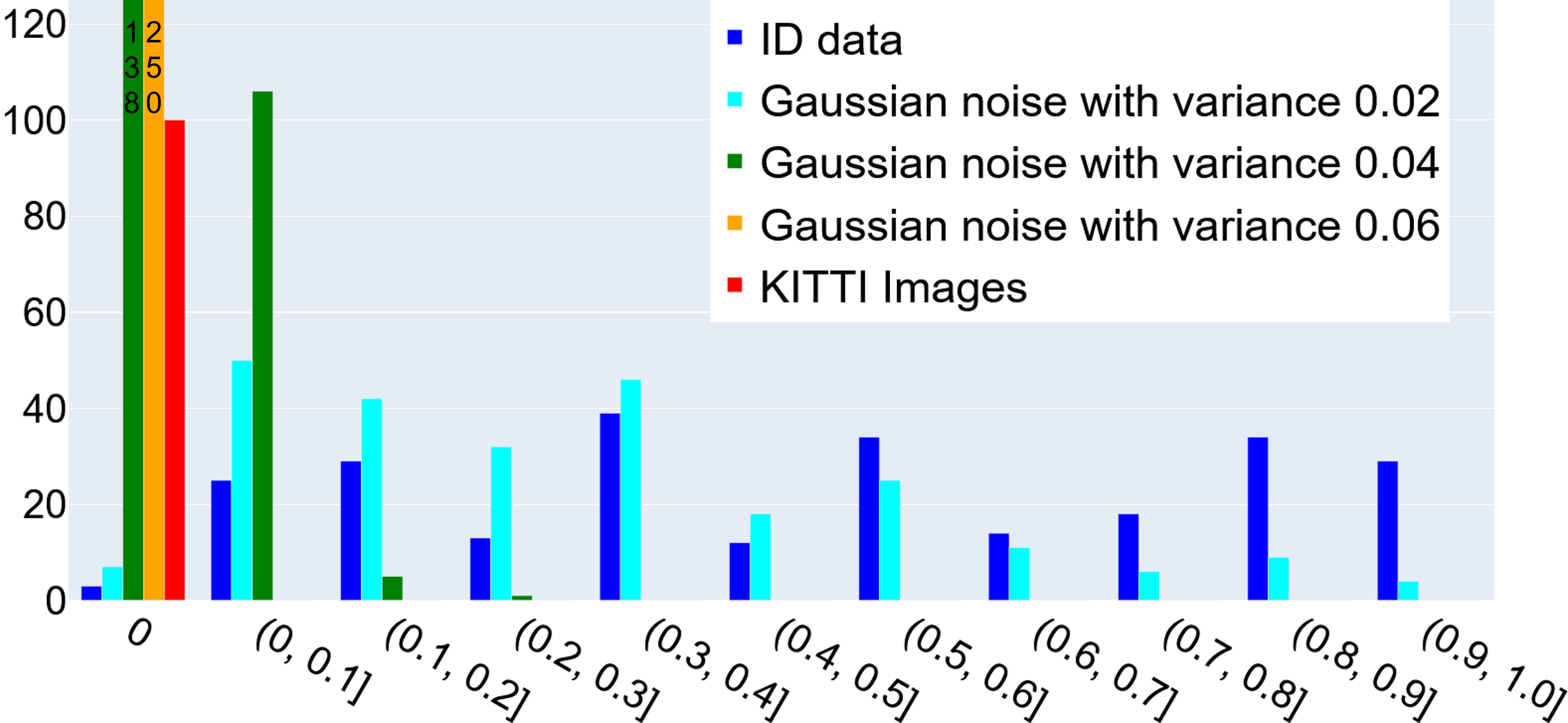}
			\caption{Last batch normalization layer}
			\label{fig:sfig1conv}
		\end{subfigure}
	}
	\caption{Number of images with the respective p-value. The x-axis shows the p-value, the y-axis the number of images resulting in the specific p-value.}
	\label{fig:histogram}
\end{figure}

\begin{figure}
	\makebox[\linewidth][c]{
		\begin{subfigure}{.3\textwidth}
			\centering
			\includegraphics[width=.9\linewidth]{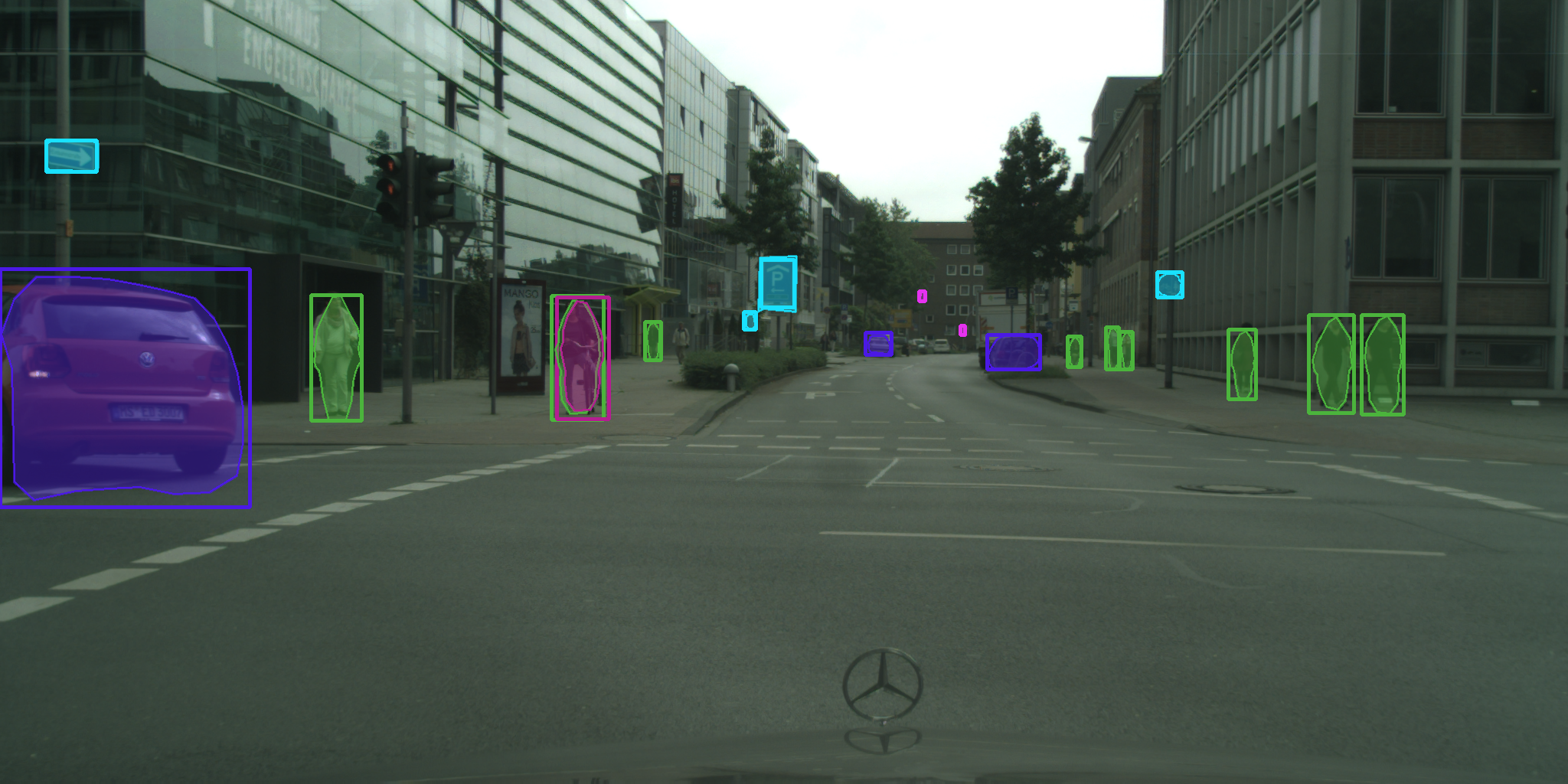}
			\small{\caption{Original image\newline}}
			\label{fig:sOrigNoise}
		\end{subfigure}%
		\begin{subfigure}{.3\textwidth}
			\centering
			\includegraphics[width=.9\linewidth]{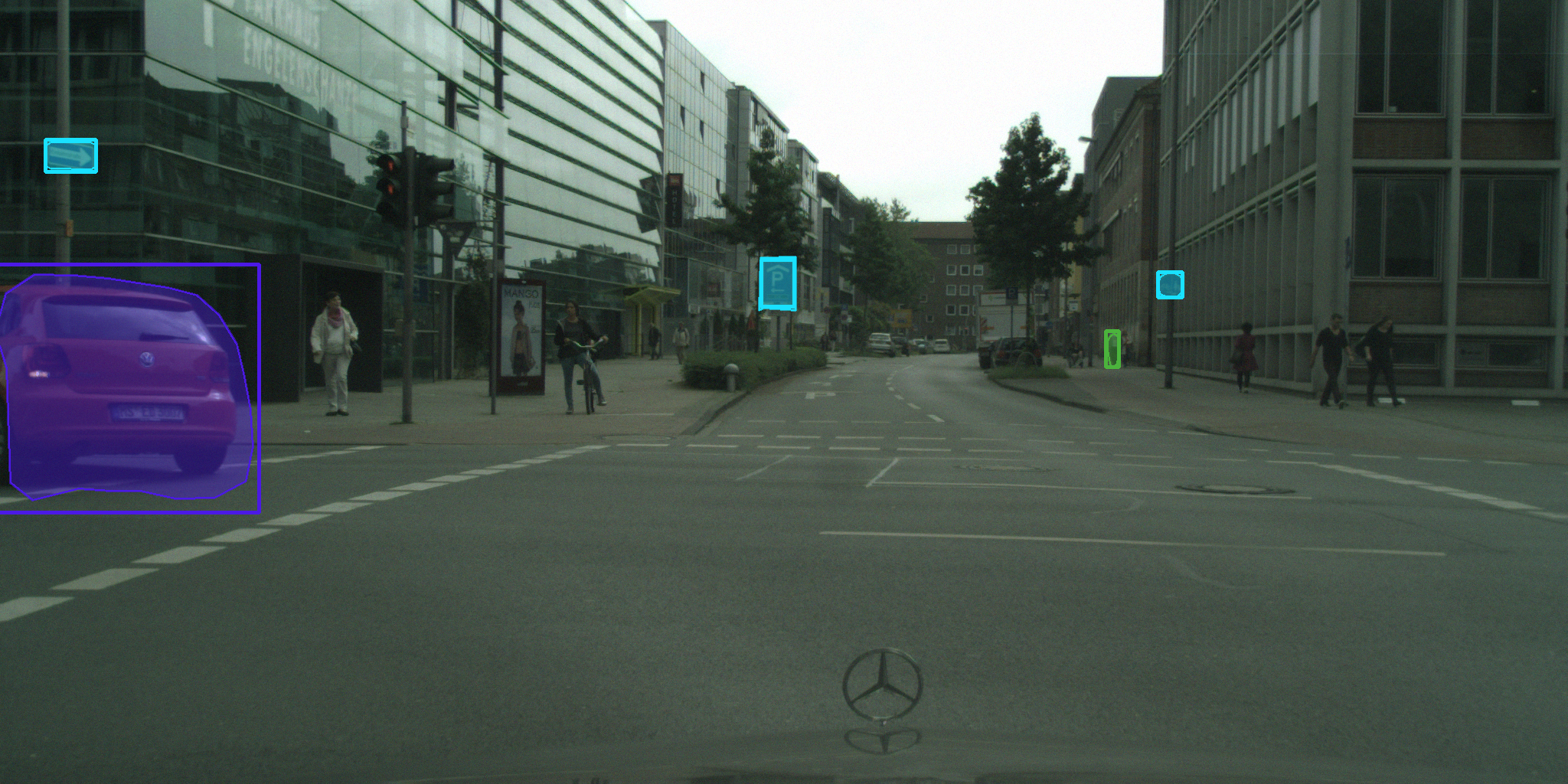}
			\caption{Gaussian Noise with variance $0.02$}
			\label{fig:s002Noise}
		\end{subfigure}
		\begin{subfigure}{.3\textwidth}
			\centering
			\includegraphics[width=.9\linewidth]{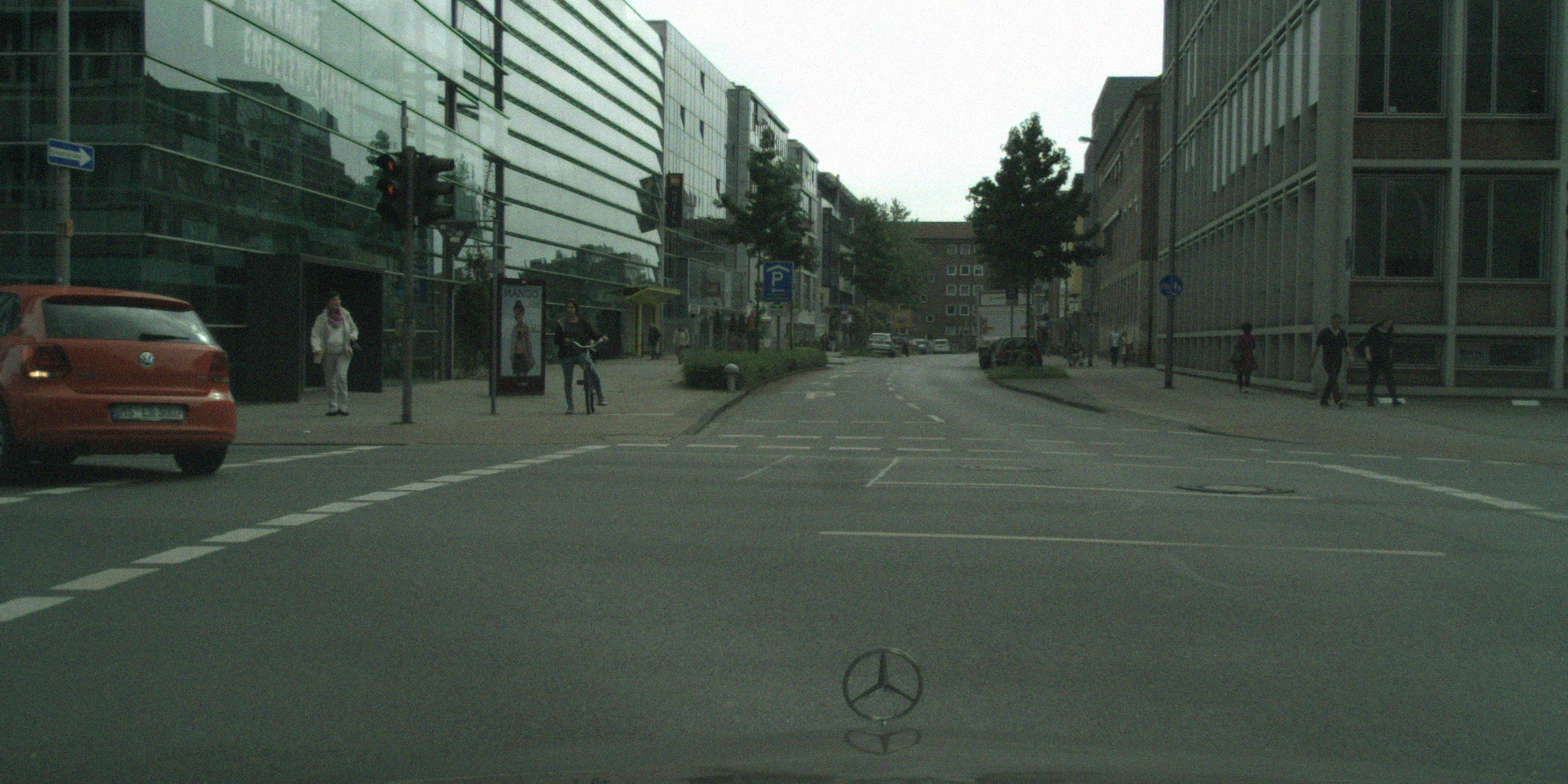}
			\caption{Gaussian Noise with variance $0.04$}
			\label{fig:s004Noise}
		\end{subfigure}
	}
	\makebox[\textwidth][c]{
		\begin{subfigure}{.3\textwidth}
			\centering
			\includegraphics[width=.9\linewidth]{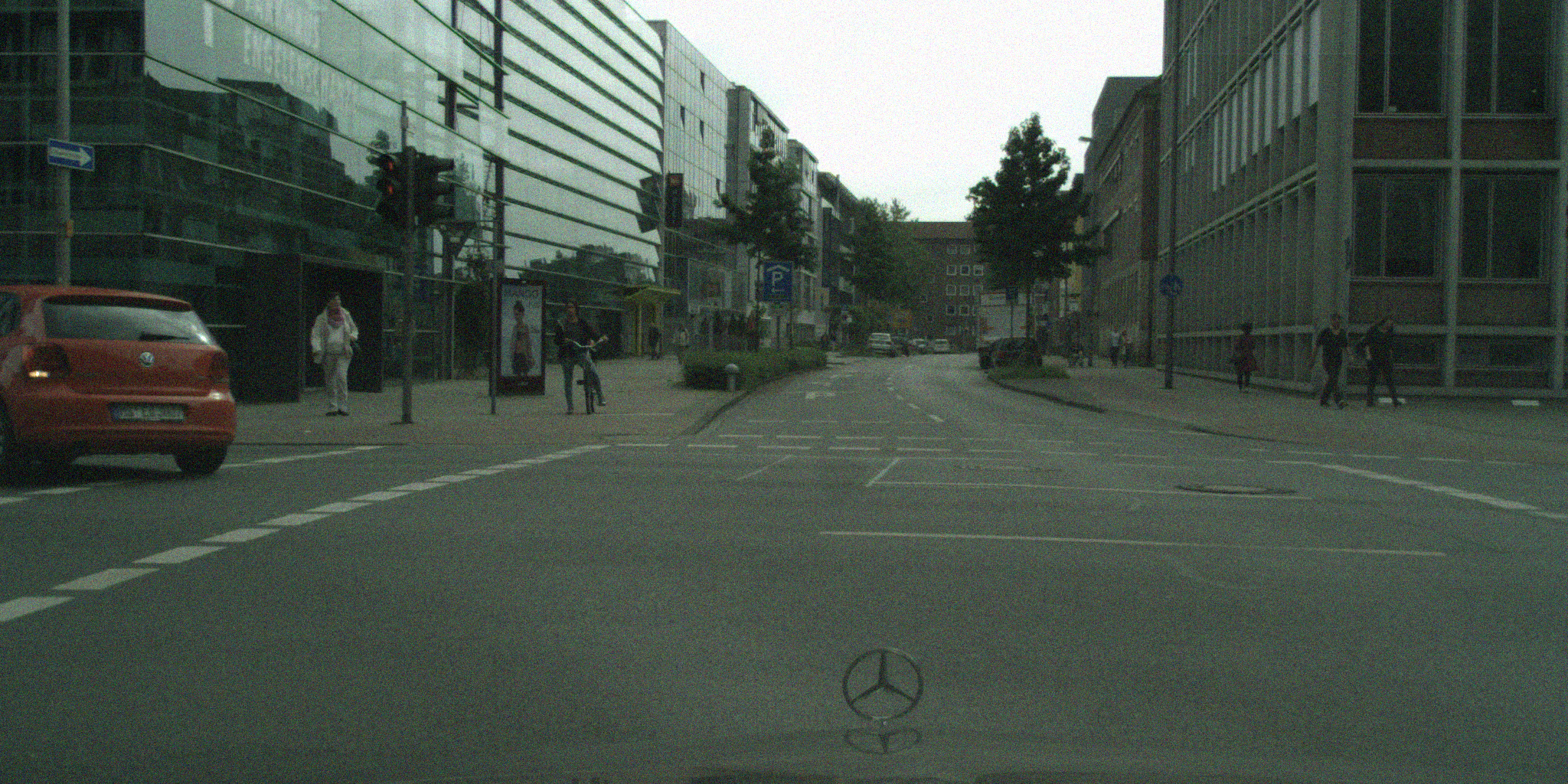}
			\caption{Gaussian Noise with variance $0.06$}
			\label{fig:s006Noise}
		\end{subfigure}
		\begin{subfigure}{.3\textwidth}
			\centering
			\includegraphics[width=.9\linewidth]{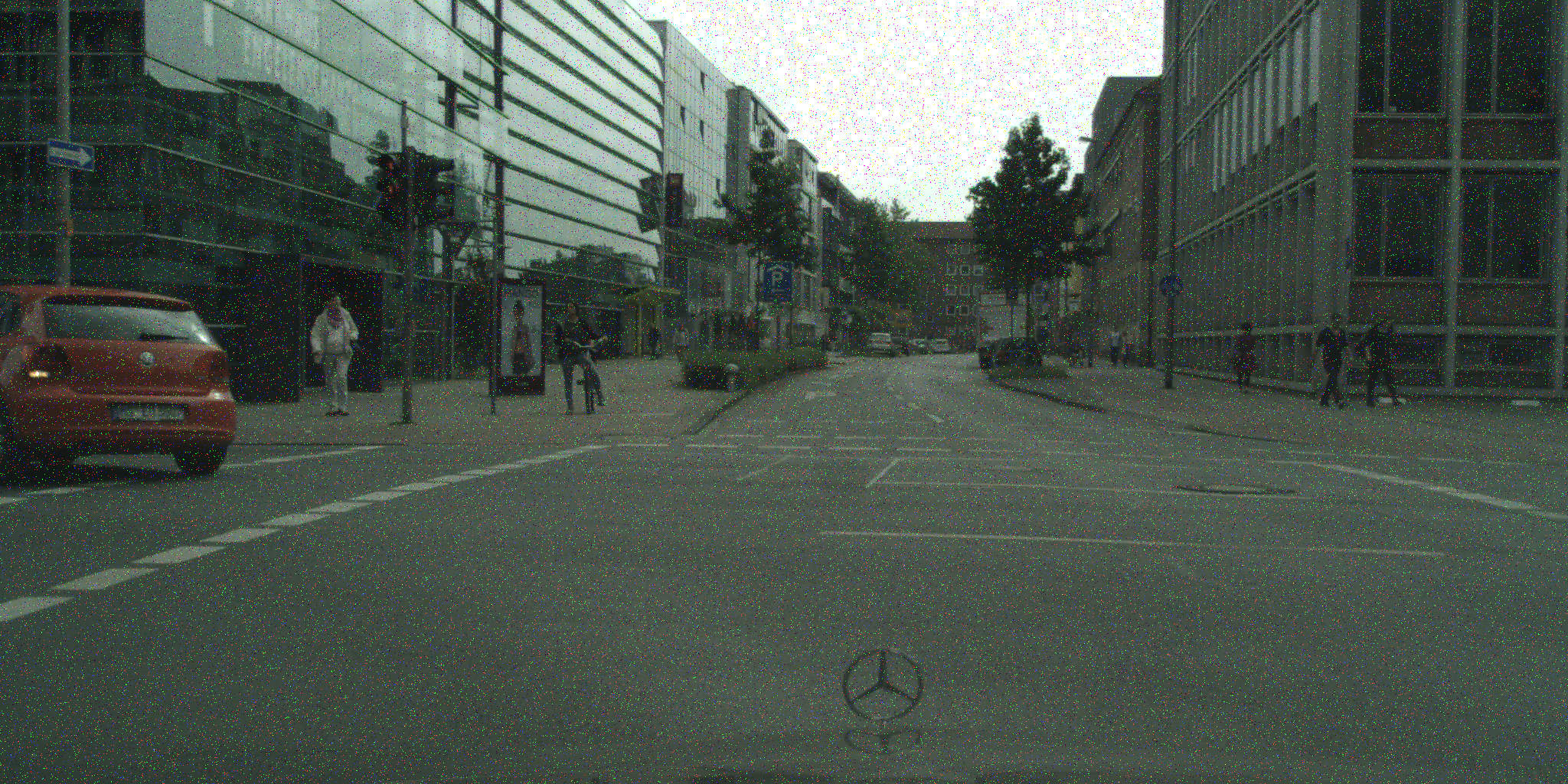}
			\caption{Impulse Noise with random parameter $0.03$}
			\label{fig:shot1}
		\end{subfigure}
		\begin{subfigure}{.3\textwidth}
			\centering
			\includegraphics[width=.9\linewidth]{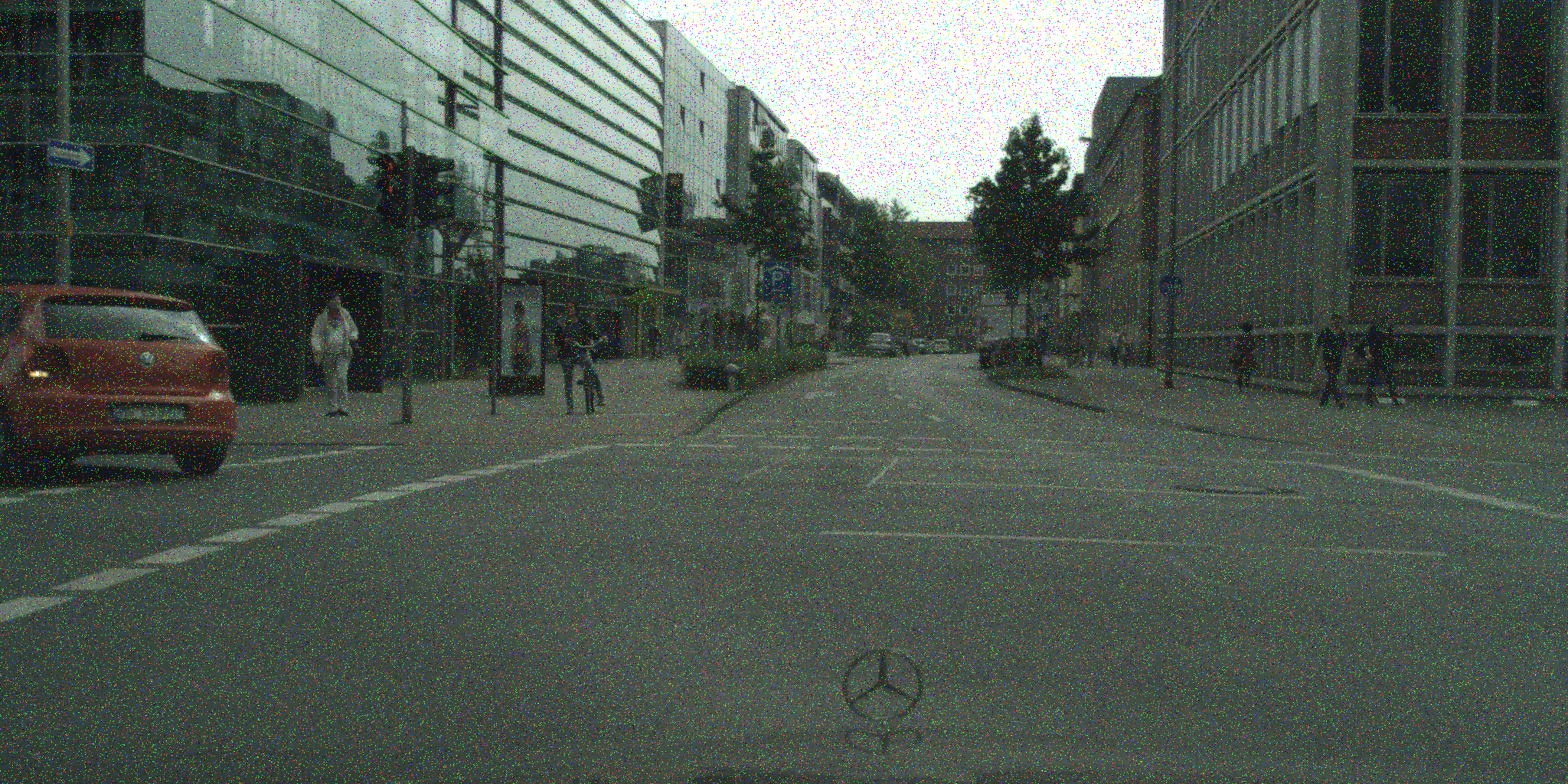}
			\caption{Impulse Noise with random parameter $0.06$}
			\label{fig:shot2}
		\end{subfigure}
	}
	\makebox[\linewidth][c]{
		\begin{subfigure}{.3\textwidth}
			\centering
			\includegraphics[width=.9\linewidth]{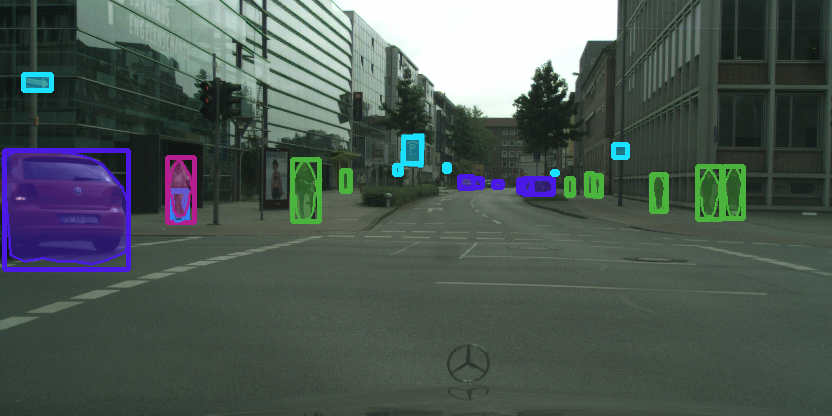}
			\caption{FGSM Attack with mask of $0.02$ times attack}
			\label{fig:fgsm2}
		\end{subfigure}
		\begin{subfigure}{.3\textwidth}
			\centering
			\includegraphics[width=.9\linewidth]{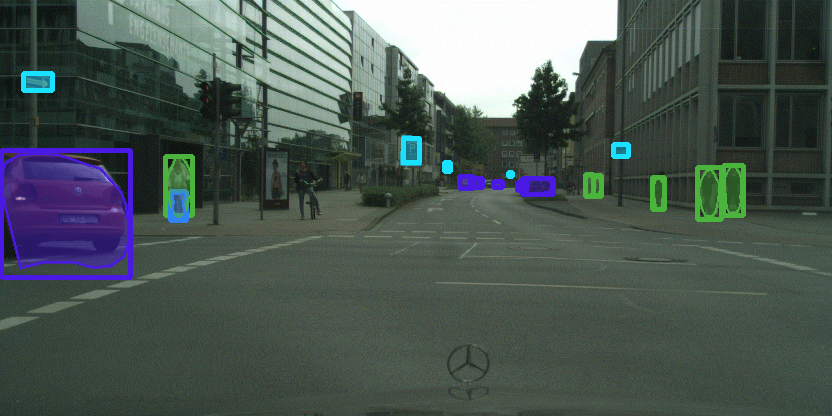}
			\caption{FGSM Attack with mask of $0.04$ times attack}
			\label{fig:fgsm4}
		\end{subfigure}
		\begin{subfigure}{.3\textwidth}
			\centering
			\includegraphics[width=.9\linewidth]{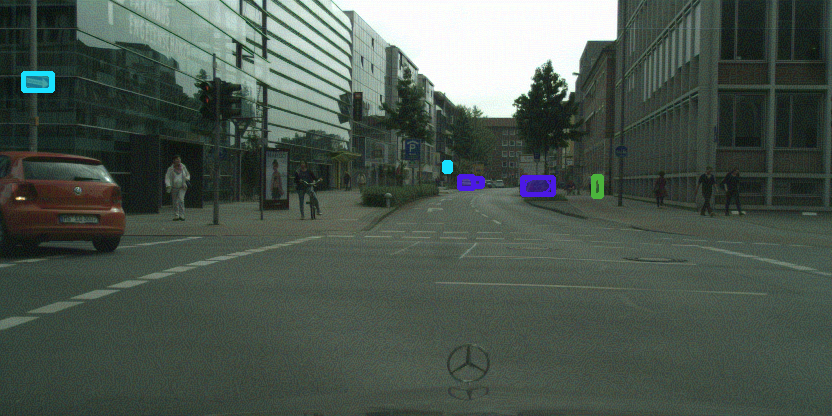}
			\caption{FGSM Attack with mask of $0.06$ times attack}
			\label{fig:fgsm6}
		\end{subfigure}
	}
	\caption{Image from the Cityscapes data set with additional perturbations and the predictions obtained from PolyYolo on the perturbed image}
	\label{fig:noiseImage}
\end{figure}

Figure \ref{fig:histogram} then shows the behavior of the p-values on selected \ac{ood} data in comparison to ID data. The x-axis represents again the intervals of the p-values while the y-axis shows the number of images with p-values ranging in the specified interval. The blue bars represent 250 ID images obtained from the validation set of Cityscapes. The respective p-values are visualized with blue color. Similarly to the setting of Hashemi et al. \cite{gaussbasedvahid} we obtained \ac{ood} data by using a different data set, namely KITTI \cite{kitti}, which also contains images captured by a vehicle driving in a German city. However, all randomly selected 100 images from the KITTI data set resulted in a p-value of $0$ which is indicated with the red bar. Therefore, we generated \ac{ood} examples from the 250 Cityscapes images we used as test data by adding Gaussian noise, as noise can be used to fool a neural network \cite{Dodge2017,Hosseini2017,noiseattackimagesegmentation}. Our implementation is based on \cite{corruptionGit}. We considered additional Gaussian noise with mean $0$ and variance $0.02$, $0.04$ or $0.06$. The noise is barely detectable for humans (see Figure \ref{fig:noiseImage}) but leads to sever faults in PolyYolo. As indicated in Figure \ref{fig:noiseImage}, a noise of variance $0.02$ already leads to a huge decrease in detection rate and for  a larger variance no objects were detected correctly.  
In Figure \ref{fig:histogram} the behavior of the p-values for images with additional noise is portrayed. The noises of variance $0.02$, $0.04$ and $0.06$ are depicted by cyan, green and orange bars, respectively. 
For better readability, some bars were shortened. It can be seen that the p-values decrease when the severity of the noise increases. This trade off can be considered when selecting a threshold value at runtime in order to decide when to raise an alarm. 

For the evaluation of the monitor in a practical setting we set the threshold for p-values to $5\%$ meaning that a sample is classified as ID if it has a higher p-value than at least $5\%$ of the calibration set. This decision was influenced by Figure \ref{fig:histogram}. Most samples perturbed with a severe Gaussian noise and  only a small portion of ID are classified as \ac{ood} by this threshold. The experiments were carried out on 100 previously unseen images of the Cityscapes data set as well as 100 images of KITTI and A2D2 \cite{a2d2}. Perturbations were applied to the Cityscapes images. In addition to Gaussian noise we used impulse noise, also called salt-and-pepper noise, and the \ac{fgsm} attack\cite{fgsm}. The impulse noise manifests as white and black pixels in the image and the strength is influenced by the random parameter. Our implementation is again based on \cite{corruptionGit}. The FGSM attack corrupts the input pixels based on the gradient of the output. The gradient is used to calculate a mask of changes which is then added to the input image. The mask is usually multiplied with a small factor to make the attack less obvious to humans. Examples of the perturbations can be seen in Figure \ref{fig:noiseImage}. 

Results of the experiment are shown in Table \ref{table_KPIs}. The number of ID data classified as OOD data lies within the range of expected values due to the setting of the threshold to $5\%$. Both layers detect Gaussian noise with variance of $0.04$ and $0.06$ while a variance of $0.02$ can fool the approach. However, this noise is not as critical as large objects are still detected from the network (see Figure \ref{fig:noiseImage} for an example). For the attacked images, the leaky ReLU layer was more precise. This is presumably due to the fact that in the FGSM images pixels were purposely changed to make a large impact on the output of the network. The leaky ReLU layer is a successor of the batch normalization layer and the last layer before the output layer. Therefore, the changes should reflect more. Furthermore, it is noticeable that all images taken from different data sets were classified correctly. 

\begin{table}[!htb]
	
	\setlength{\tabcolsep}{6pt}
	
	\centering
	\caption{The table shows the number of images classified as ID and OOD dependent on the perturbation applied and the data set used. Noise and FGSM were applied to the ID data.}
	
	\begin{tabulary}{\textwidth}{LCCCC}
		
		\toprule
		
		\multirow{2}{*}[-0.25em]{Data} & \multicolumn{2}{c}{Leaky ReLU layer} & \multicolumn{2}{c}{Batch normalization layer}\\
		
		\cmidrule(lr){2-3} \cmidrule(lr){4-5} \\
		
		\addlinespace[-8pt]
		
		{} &  Classified as ID & Classified as OOD &  Classified as ID &  Classified as OOD  \\
		
		\midrule
		
		ID data & 97 &3 & 94 & 6\\ \addlinespace[4pt]
		Gaussian noise with variance 0.02 & 93 & 7 & 91 & 9 \\ \addlinespace[4pt]
		Gaussian noise with variance 0.04 & 9 &91 & 8& 92 \\ \addlinespace[4pt]
		Gaussian noise with variance 0.06 & 0 &100 & 0 & 100\\ \addlinespace[4pt]
		Impulse noise with random parameter 0.03 & 0 &100 & 0 & 100\\ \addlinespace[4pt]
		Impulse noise with random parameter 0.06 & 0 &100 & 0 & 100 \\ \addlinespace[4pt]
		FGSM with mask multiplied by 0.02 & 35 &65 & 39 & 61\\ \addlinespace[4pt]
		FGSM  with mask multiplied by 0.04& 8 &92 & 11 & 89\\ \addlinespace[4pt]
		FGSM  with mask multiplied by 0.06 & 0 &100 & 0 & 100\\ \addlinespace[4pt]
		KITTI & 0 & 100  & 0 & 100\\ \addlinespace[4pt]
		A2D2 & 0 & 100 & 0 & 100\\
		
		\bottomrule
		
	\end{tabulary}
	
	\label{table_KPIs}
	
\end{table}

\section{Conclusion and Future Work}
In this work we developed a tool to detect OOD images at runtime for 2D object detection systems. The idea was based on Gaussian monitoring of the neuron activation patterns. We additionally embedded the method into the framework of inductive conformal anomaly detection to receive a quantitative measure of difference between the training set and new samples. Experiments visualizing the p-values were carried out. 

The proposed idea can be extended in several ways. First of all, the selection of images for the calibration set can be improved as we observed a difference in importance for the randomly selected images. In addition, the selection of monitored layers requires further evaluation. We only considered the last two hidden layers of the network. However, the architecture of PolyYolo contains staircase upsampling with skip connections. Activation values obtained from these connections are a natural way to extend the monitoring approach to also take intermediate neuron values into consideration. Furthermore, more experiments on other neural network architectures are required in order to generalize the results. For the same reason, different types of perturbations and attacks should be considered for generating \ac{ood} data. An extension of the MaxLogit monitor from \cite{Hendrycks2019} to the application of object detection with the goal of comparing both monitors is worth to be exploited.

\bibliographystyle{splncs04}
\bibliography{bibliography}

\end{document}